\documentclass[11pt, a4paper]{article}
\usepackage[utf8]{inputenc}
\usepackage[a4paper, margin=1in]{geometry}
\usepackage{epsfig} 
\usepackage{mathptmx} 
\usepackage{times} 
\usepackage{amsmath,amssymb,amsfonts} 
\usepackage{cite}
\usepackage{graphicx}
\usepackage{textcomp}
\usepackage{wrapfig}
\usepackage{color}
\usepackage{bm}
\usepackage{multirow}
\usepackage{booktabs}
\usepackage{array}
\usepackage{gensymb}
\usepackage{tabularx}
\usepackage{float}
\usepackage{microtype}
                                    
\title{\LARGE \bf
Tactile Image-to-Image Disentanglement of Contact~Geometry from Motion-Induced Shear 
}
\author{
  Anupam K.~Gupta\\
  Department of Engineering Maths and Bristol Robotics Laboratory\\
  University of Bristol
  United Kingdom\\
  \texttt{anupam.gupta@bristol.ac.uk} \\
  Laurence Aitchison \\
  Department of Computer Science \\
  University of Bristol 
  United Kingdom\\
  \texttt{laurence.aitchison@bristol.ac.uk} \\
  Nathan F. Lepora \\
  Department of Engineering Maths and Bristol Robotics Laboratory\\
  University of Bristol
  United Kingdom\\
  \texttt{n.lepora@bristol.ac.uk} \\
}

\begin{document}
\maketitle
\thispagestyle{empty}
\pagestyle{empty}

\begin{abstract}
Robotic touch, particularly when using soft optical tactile sensors, suffers from distortion caused by motion-dependent shear. The manner in which the sensor contacts a stimulus is entangled with the tactile information about the geometry of the stimulus. In this work, we propose a supervised convolutional deep neural network model that learns to disentangle, in the latent space, the components of sensor deformations caused by contact geometry from those due to sliding-induced shear. The approach is validated by reconstructing unsheared tactile images from sheared images and showing they match unsheared tactile images collected with no sliding motion. In addition, the unsheared tactile images give a faithful reconstruction of the contact geometry that is not possible from the sheared data, and robust estimation of the contact pose that can be used for servo control sliding around various 2D shapes. Finally, the contact geometry reconstruction in conjunction with servo control sliding were used for faithful full object reconstruction of various 2D shapes. The methods have broad applicability to deep learning models for robots with a shear-sensitive sense of touch.
\end{abstract}


\vspace{-0.5em}
\begin{figure}[h]
  \centering
  \includegraphics[width=1\textwidth]{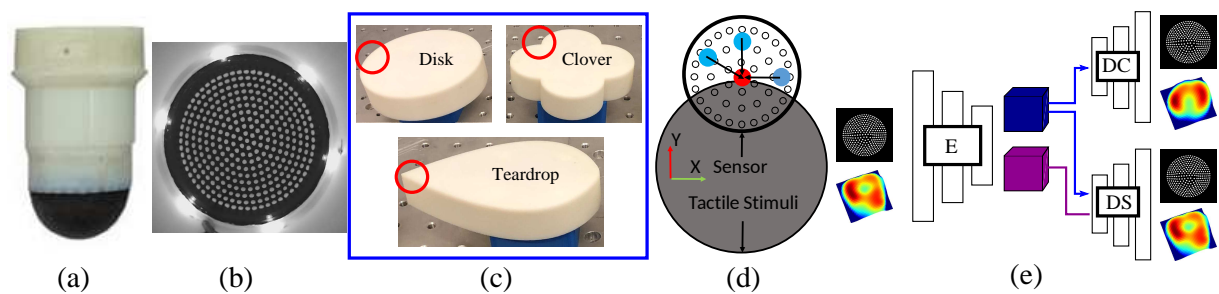}
  \vspace{-1.5em}
  \caption{\textbf{Experiment setup and model architecture.} (a) TacTip sensor, (b) Sensor internal surface, (c) 3D-printed tactile stimuli with approximate locations of data collection (red circles), (d) Schematic of data collection, showing target location (red dot) and initial contact location (blue dots). Canonical (tapping) data was collected by vertically contacting the target; sheared (sliding) data was collected by sliding to the target, and (e) Schematic of the \textit{Disentangled UnshearNet} model architecture with representative input/output tactile images and contact reconstructions.}
  \vspace{-0.5em}
  \label{fig:sensor}
\end{figure}

\vspace{-0.5em}
\section{Introduction}
\vspace{-0.5em}

A fundamental aspect of tactile sensing is that it measures the environment via physical contact. Therefore, the manner in which the sensor contacts a stimulus is entangled with the tactile information about that stimulus. For example, when sliding or rubbing fingers across tactile stimuli, soft tactile sensors are inevitably distorted by motion-dependent shear, making the sensor response history-dependent and generalization hard in tactile tasks. This issue is prevalent in camera-based optical tactile sensors that use the lateral motion of markers to infer contact geometry, such as the TacTip~\cite{Ben_Tactip}, ChromaTouch~\cite{Sensor_ChromaTouch}, GelForce~\cite{kamiyama_vision-based_2005} and the optical tactile sensor described in~\cite{Sensor_Andrea}, but may occur also in other shear-sensitive tactile sensors. Generally, this shear distortion of the sensor response when sliding along surfaces would make it difficult to perform tactile-dependent tasks such as shape reconstruction and continuous object exploration or manipulation.  

Therefore, an important capability is to separate the deformation due to the manner of contact from that due to the stimulus geometry. For example, when continuously exploring an object surface, the sliding-induced shear would need to be removed from tactile image to extract the local contact shape. Previous work on extracting pose information from contact to continuously slide a tactile sensor over complex objects used a deep convolutional neural network trained to map between shear-distorted inputs and the desired (pose) outputs, which was necessary for sliding exploration over a range of complex novel objects~\cite{Nathan_CNN_Sliding1, Nathan_CNN_Sliding2, Nathan_CNN_Sliding3}. Here, we suggest, an alternate  disentanglement based approach that learns to remove motion-induced shear from  raw sheared sensor response that has several advantages over previous task-specific approach of training insensitivity to shear, in that: (a)~it is computationally and data efficient by only requiring the ‘unshearing’ of the tactile data to be learnt once instead of relearning for each tactile task, such as when applied to full object reconstruction; (b)~the models can then be learnt more easily on unsheared data, which we demonstrate by predicting local object pose; moreover, in some cases no model need be learnt at all, which we demonstrate with contact surface reconstruction using a Voronoi tessellation from the tactile marker data; and (c) the methods are extensible to including other downstream variables of interest appropriate to the task, rather than having to completely retrain new models from scratch. 

Our approach to disentangling environmental factors of variation and object attributes in tactile sensing is based on work in computer vision where disentanglement is considered important to improve the generalizability of machine learning models, for example to aid in learning of robust object representations for object classification under novel scenarios~\cite{bengio_disentanglement}. Inspired from this, here we propose a supervised convolutional neural network model that learns to disentangle the components of sensor deformation caused by the contacted object geometry from those due to motion-induced shear. The models were trained using paired data from discrete (tapping) and continuous (sliding) touch, with an image-to-image `Disentangled UnshearNet' model architecture that outputs unsheared tactile images from sheared images  (Fig.~\ref{fig:sensor}). Our validation: (1) quantifies that the unsheared images from continuous motion match their paired tactile images from discrete taps (Table~\ref{tab:ssim}); (2)~demonstrates that local contact geometry reconstructed from unsheared images using Voronoi match their paired tapping counterparts over several stimuli shapes (Fig.~\ref{fig:results}, top row); (3) infers pose from the sheared tactile images by re-using a convolutional neural network model trained on discrete (tapping) data, which was then applied to controlling robust sliding exploration around several planar objects (Fig.~\ref{fig:results}, middle row); and (4) combines the two previous steps to attain faithful full-object reconstruction of several planar shapes (Fig.~\ref{fig:results}, bottom row). 

\vspace{-0.5em}
\section{Background}
\vspace{-0.5em}

Measuring strain is a fundamental component of tactile sensing, as emphasised by Platkiewicz, {\em et al}~\cite{platkiewicz_haptic_2016}: `the pressure distribution at the surface of a tactile sensor cannot be acquired directly and must be
inferred from the [strain] field induced by the touched object in the sensor medium.' 

There are many mechanisms whereby normal strain and shear strain are measured by tactile sensors. Taxel-based tactile sensors such as the iCub fingertip~\cite{wettels_biomimetic_2008} and BioTac~\cite{schmitz_methods_2011} directly measure normal strain. Nevertheless, this measurement of normal strain will be affected by shearing of the soft sensor medium, although the effects on sensor output will be much smaller than those from normal pressure.   


Camera-based optical tactile sensors can be very sensitive to shear because they transduce deformation of the soft sensor surface into a lateral image. For optical tactile sensors such as the TacTip biomimetic optical sensor used here~\cite{Rossiter_TactipOrg,Ben_Tactip} and other marker-based sensors~\cite{Sensor_ChromaTouch,Sensor_Andrea}, the surface shear will also distort the apparent geometry of the contact. This distortion is because the normal indentation is represented in the local shear strain of the markers, upon which the global shear strain is superimposed. This contrasts with optical tactile sensors that reconstruct the indentation field from reflection, such as the GelSight~\cite{Adelson_GelSight}, where shear will shift the imprint on the tactile image but otherwise the imprint shape should be relatively unaffected. 


This effect of shear complicates tactile perception under general motions against an object, and restricts the application of touch on challenging tasks such as tactile exploration and surface shape reconstruction. Studies have been limited to discrete contacts where the sensor taps discretely over the object to minimize shear~\cite{Nathan_Tap_ContourFollow1, Nathan_Tap_ContourFollow2, ShanLuo_Tap_RL_Gelsight_SurfaceFollow}, or to rigid planar tactile sensors~\cite{Li_Sliding_SurfaceFollow_NonDeformableSensor, Kappassov_Sliding_SurfaceFollow_NonDeformableSensor} that do not shear. 

A recent body of work on the TacTip sensor used here focused on mitigating the effects of motion-dependent shear during tactile servo control to slide around planar shapes~\cite{Nathan_CNN_Sliding1, Kirsty_ShearInv_PCA} and complex 3D objects~\cite{Nathan_CNN_Sliding2,Nathan_CNN_Sliding3}. The shear-dependence problem was addressed in three studies that used a deep convolutional neural network to predict pose by training insensitivity to shear into the prediction network~\cite{Nathan_CNN_Sliding1,Nathan_CNN_Sliding2,Nathan_CNN_Sliding3}. While effective, this has a drawback that procedures for learning shear-invariance needs to be introduced into the training for each individual task, and thus would not generalize easily to new tasks. Another study did not use deep learning, but instead dimensionally reduced the tactile data~\cite{Kirsty_ShearInv_PCA}, where the the pose components naturally separated from the shear components. 

This work pursues an alternate approach using deep learning to remove the effect of motion-dependent shear from the sheared tactile image, offering several advantages that were summarized in the introduction. Our approach to disentangling environmental factors of variation and object attributes in tactile sensing is based on work in computer vision where disentanglement is considered important to improve the generalization of machine learning models, for example to aid in learning of robust object representations for object classification under novel scenarios~\cite{bengio_disentanglement}. Once disentangled, different environmental factors and object attributes can be recombined to generate coherent novel concepts thus extending knowledge to previously unobserved scenarios~\cite{bengio_disentanglement}. For example, synthesizing sensor responses to novel stimuli by combining the stimuli attributes and factor of variation without actually observing them to plan action under previously unobserved conditions. In the case of tactile sensing considered here, the stimuli attributes could be novel contact geometry and the factors of variation the motion-induced shear.

\vspace{-0.5em}
\section{Methods}
\vspace{-0.75em}
\subsection{Experimental Setup}
\vspace{-0.5em}
The setup in this study followed recent experiments with the TacTip biomimetic optical tactile sensor \cite{Nathan_CNN_Sliding1, Kirsty_ShearInv_PCA,Nathan_CNN_Sliding2,Nathan_CNN_Sliding3} where it is mounted as an end effector on a robot arm (Fig.~\ref{fig:sensor}). The TacTip is biomimetic in that its skin morphology is based on the layered dermal and epidermal structure of the human fingertip~\cite{Rossiter_TactipOrg,Ben_Tactip}; the tactile sensor measures the shear displacement of an array of papillae pins on the inner side of the sensing surface caused by the skin deformation.  

The TacTip can be designed with varied morphologies dictated by the application~\cite{Ben_Tactip}. The TacTip sensor (Fig.~\ref{fig:sensor} (a) \& (b)) used in this work had a soft rubber-like hemispherical dome (40\,mm radius, TangoBlack+) manufactured using multi-material 3D printing. Its inner side was covered with an array of 331 pins with hard white tips arranged in a concentric circular grid. The dome was filled with an optically-clear silicone gel to give it a compliance similar to a human fingertip. The final assembly of sensors is done by hand that introduces variability in the response in other otherwise same sensors. A camera was used to collect tactile images of the internal surface of the dome. For more details, we refer to the references above.


The tactile stimuli used here were three 3D-printed (ABS plastic) planar shapes: a circular disk, clover and teardrop (Fig.~\ref{fig:sensor} (c)). These objects were chosen to give different edge shapes for application of our methods, and were fastened securely to the workspace to prevent accidental motion. An additional four complex planar shapes, including two acrylic shapes with distinct frictional properties to the remaining 3D-printed shapes (spiral 1 \& 2, Fig.~\ref{fig:results}) were used to test for generalizability.

  
\vspace{-0.5em}
\subsection{Data Collection}
\vspace{-0.5em}
\subsubsection{Paired Canonical and Sheared Data}
\vspace{-0.5em}
This work proposes a supervised deep learning model \textit{Disentangled UnshearNet} to remove the effect of motion-induced global shear in the tactile images, caused by the friction between the sensor skin and its contacted surface during contact. We emphasise this motion-induced shear is different from the skin deformation due to local shear from the geometry of the stimulus imprint. 

Our method requires paired tactile data with minimal global shear taken from vertically tapping onto the stimulus (referred to as canonical data) and sliding data with random global shear (referred to as sheared data). The sheared data is collected after sliding randomly and is paired with a tactile image from the canonical data by using the same relative poses between the sensor and the stimuli (Fig.~\ref{fig:sensor}). The raw tactile data was cropped, sub-sampled to a $256\times256$-pixel region and finally converted to binary images using an adaptive threshold to minimise sensitivity to internal lighting conditions. 



In total, data for 200 distinct poses were collected for each of the three stimuli (Fig.~\ref{fig:sensor} (c)). To generate poses, the sensor location relative to the target contact location (red circle, Fig.~\ref{fig:sensor} (d)) was sampled randomly from a uniform distribution spanning the range [-5, 5]\,mm in the two lateral directions (along the $x$- and $y$-axes), [-45, -45]\,deg in yaw ($\theta n_{Z}$) and [-6, -1]\,mm in depth ($z$-axis). Ranges of the target poses and perturbations to generate canonical and sheared data were chosen to ensure safe contact of the sensor with the stimulus (Fig.~\ref{fig:sensor} (c)).

\vspace{-0.5em}
\paragraph{\textbf{Canonical Data:}}
The canonical data was taken by vertically tapping onto the stimulus, to give a representative sample of tactile data with minimal global shear just due to the stimulus geometry. This gives target (reference) data that the sheared data from sliding contacts should be restored to. The dataset, in total, had 30,000 samples: 50 instances of each of the 200 poses recorded previously for each of the three stimuli shapes (Fig.~\ref{fig:sensor} (c)). To generate instances, the indentation depth ($z$-axes) was varied randomly between [-1, 1]\,mm from the original indentation depth.  

\begin{figure*}[ht]
  \centering
  \includegraphics[width=0.9\textwidth]{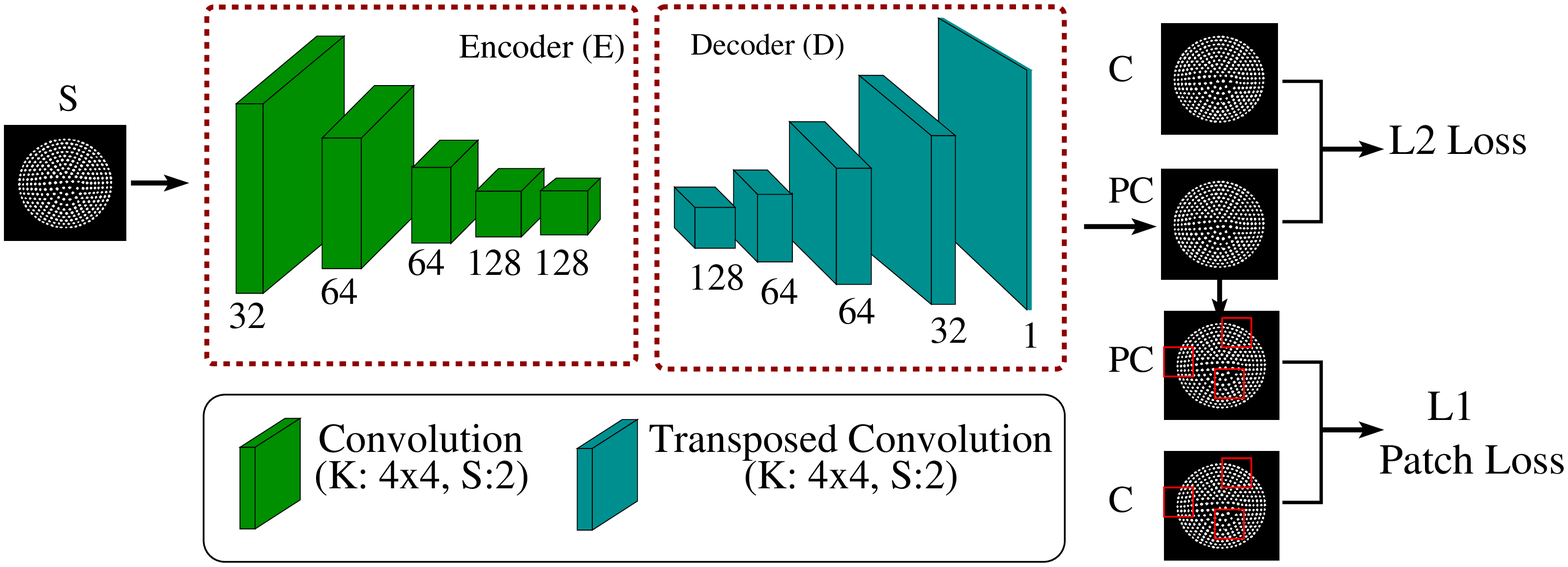}
  \vspace{-.5em}
  \caption{\textbf{Entangled UnshearNet architecture:} A shear-distorted tactile image \textit{S} was processed by the encoder \textit{E} and decoder \textit{D} (kernel \textit{K}, stride \textit{S}) with output \textit{PC}: the unsheared counterpart of input \textit{S} post removal of sliding-induced distortion. The match between predicted unsheared data \textit{PC} and canonical data~\textit{C} was enforced by training with a direct comparison between \textit{PC} and \textit{C} along with comparing corresponding patches ($20\times20$ pix.) taken from \textit{PC} and \textit{C} to enforce local compliance.}
  \label{fig:base_arch}
  \vspace{1em}
  \includegraphics[width=0.9\textwidth]{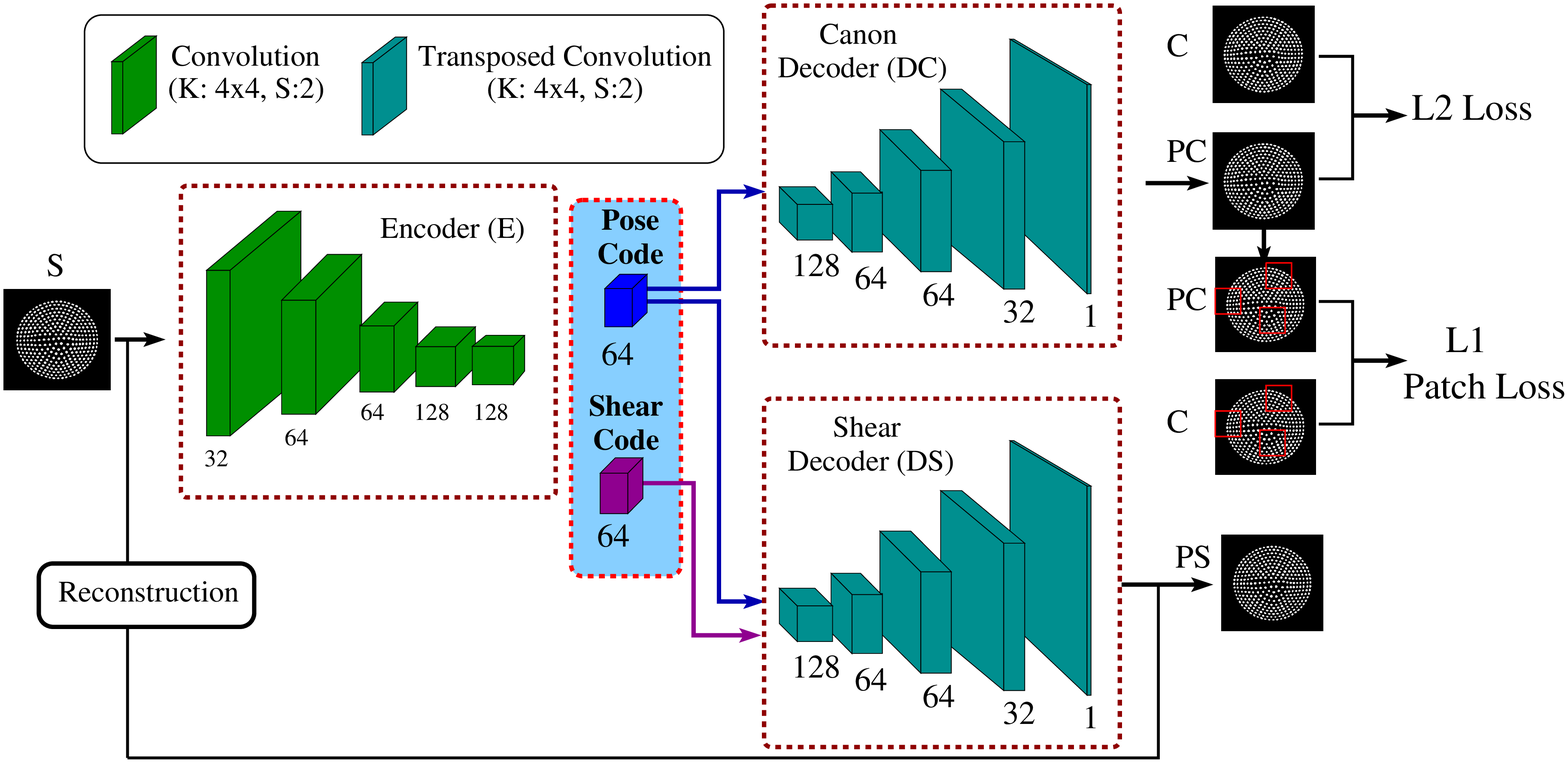}
  \vspace{.5em}
  \caption{\textbf{Disentangled UnshearNet architecture:} Labelling as in Fig.~\ref{fig:base_arch}, with two decoders \textit{DC} and \textit{DS}. The first decoder has output \textit{PC}: the unsheared counterpart of input $S$ post removal of sliding-induced distortion; the second decoder \textit{DS} reconstructed the input \textit{S}, denoted \textit{PS}.}
  \label{fig:prop_arch}
\end{figure*}

\vspace{-0.5em}
\paragraph{\textbf{Sheared Data:}}
The sheared data accords to a canonical tactile image distorted by global shear from lateral sliding motion of the sensor on the stimulus. Thus, the sheared data does not give contact-only information akin to canonical data that represented the geometry of the contacted stimulus, but has a significant component due to the contact motion. 

The dataset, in total,  had 90,000 samples: 150 instances of each of the 200 random poses recorded previously for each of the three stimulus shapes (Fig.~\ref{fig:sensor} (c)). To generate instances for each pose, the lateral position offset for sliding to the target pose was varied randomly between [-2.5, 2.5]\,mm in lateral direction(s) (along the $x$- or $y$-axes or both), using the procedure shown in Fig.~\ref{fig:sensor} (d). 
\vspace{-0.5em}
\paragraph{\textbf{Training and Test Data:}}
The canonical and sheared datasets were partitioned randomly into training, validation and test sets (ratio 60:20:20); i.e. from the 200 unique poses for each stimulus shape (Fig.~\ref{fig:sensor} (c)), 120 poses were assigned to a training set and 40 poses each to validation and testing sets. Overall the canonical and sheared training set contained 18,000 and 54,000 samples respectively, while the validation and testing sets contained 6,000 and 18,000 samples respectively. To generate paired canonical and transformed data for training, each transformed tactile image was paired randomly with one of the canonical instances having the same pose. 


\vspace{-0.6em}
\subsubsection{Pose-labelled Data}
\vspace{-0.5em}
To train pose prediction network \textit{PoseNet}, a separate set of pose data was collected instead of reusing the previously collected pose data to train \textit{Entangled \& Disentagled UnshearNet}. This was done to not overfit the training set and to ensure performance of the trained \textit{UnshearNet} models despite potential dataset shifts due to a host of factors, including drift in sensor characteristics over time or with continued use due to wear and tear of the tactile sensing surface.

The dataset had, in total, 30,000 samples: 10,000 samples from each of the three object shapes (Fig.~\ref{fig:sensor} (c)). To generate contact poses, the sensor location relative to the initial contact location was sampled randomly from a uniform distribution spanning the range [-5, 5]\,mm laterally ($x$- and $y$-axis directions), [-45, -45]\,deg in yaw ($\theta n_{Z}$) and [-6, -1]\,mm in depth ($z$-axis). Ranges of the target poses, as before, were chosen to ensure safe contact of the sensor without causing damage. As discussed above, all tactile images were cropped, sub-sampled and thresholded to $256\times256$-pixel to binary (black/white) tactile images. The dataset was partitioned randomly into training , validation and test sets in the ratio 60:20:20 as above, containing 18,000, 6,000 and 6,000 samples respectively.

\vspace{-0.6em}
\subsection{Image-to-Image Models for Shear Removal}
\vspace{-0.5em}
To achieve our goal of removing the effects of shear induced by the sliding motion, we trained two models based on the encoder-decoder framework -- \textit{Entangled UnshearNet} (Fig.~\ref{fig:base_arch}) and \textit{Disentangled UnshearNet} (Fig.~\ref{fig:prop_arch}). The former model (baseline) learns the mapping between the input sheared data and its corresponding canonical (tap) data directly. The later model instead, first disentangles the contact-only (pose) sensor response component from the sliding-induced global shear component in the latent space of the encoder $E$; followed by accurate reconstruction of the input sheared image and its corresponding canonical (tap) image using decoders $DS$ (sheared or sliding) and $DC$ (canonical or tapping). We hypothesized that a more robust mapping between sheared and canonical (tap) data can be learned by disentangling the contact-only and sliding-induced components of sensor response. The mathematical formulation of the \textit{Disentangled UnshearNet} is listed below:



The encoder $E$ splits the latent space into two components $\mathbf{z=(z_{p}, z_{s})} \in Z_p\times Z_s$, given paired canonical $\{x_{c}\}_{i=1}^{N}$ and shear-transformed training samples $\{x_{s}\}_{i=1}^{N}$, where $x_{c}\in X_{c}$ and $x_{s}\in X_{s}$ are in the sets of canonical and sheared data, respectively. 

The decoder $DC$ takes as input only the pose part of the latent code $\mathbf{z}_{p}$ and reconstructs the canonical sample corresponding to the input sheared sample. In combination with the encoder $E$, it thus learns a mapping $M: X_{t} \rightarrow X_{c}$ from the sheared to canonical data. 

The other decoder $DS$, takes as input the full latent code $\mathbf{z}$ and reconstructs the input sheared sample. In combination with the encoder $E$, it thus acts as an auto-encoder and learns a mapping $N: X_{s} \rightarrow X_{s}$ from the sheared data to itself. 

Our objective function has three components:\\ 
\noindent (1) \textit{L2 loss} for matching the generated canonical samples $M(x_{t})$ to their targets $x_{c}$; \\ 
\noindent (2) \textit{L1 patch loss} for matching random patches extracted from generated canonical images $M(x_{t})$ and corresponding targets $x_{c}$ for enforcing image similarity locally as changes in sensor response due to change in contact conditions are predominately local in nature; \\
\noindent (3) \textit{Reconstruction loss} for matching the generated sheared samples $N(x_{t})$ to their target input sheared samples $x_{s}$.  

The objective functions are:
\begin{eqnarray}
    \mathbb{L}_{\rm rec}(E, DS) = \mathbb{E}_{x_{s}\sim X_{s}}
    [\|x_{t} - DS(z_{p}, z_{s})\|_{2}], \\
    \mathbb{L}_{\rm sup}(E, DC) = \mathbb{E}_{x_{s}\sim X_{s},  x_{c}\sim X_{c}} [\|x_{c} - DC(z_{p}))\|_{2}], \\
    \mathbb{L}_{\rm patch}(E, DC) = \mathbb{E}_{x_{s}\sim X_{s},  x_{c}\sim X_{c}} [\|{\rm crops}(x_{c}) - {\rm crops}(DC(z_{p})))\|_{1}],
\end{eqnarray}
where $(z_{p}, z_{s}) = E(x_{s})$ for all $x_{s}\in X_{s}$. The function ${\rm crops}()$ selects a patch of size $20\times20$\,pix from $M(x_{t})$ and $x_{c}$. The objective functions ${L}_{\rm rec}$, ${L}_{\rm sup}$ and ${L}_{\rm patch}$ represent sheared reconstruction loss, canonical supervised loss and canonical supervised patch loss respectively.

The final objectives for Encoder \textit{E} and Decoders $DC$, $DS$ are:
\begin{eqnarray}
        \mathbb{L}_{E} = \mathbb{L}_{\rm rec}(E, DS) + \mathbb{L}_{\rm sup}(E, DC) + \lambda\cdot\mathbb{L}_{\rm patch}(E, DC), \\
        \mathbb{L}_{DC} = \mathbb{L}_{\rm sup}(E, DC) + \lambda\cdot \mathbb{L}_{\rm patch}(E, DC),\\
        \mathbb{L}_{DS} = \mathbb{L}_{\rm rec}(E, DS),
\end{eqnarray}
where $\lambda$ is the relative loss scale factor.

Details of the specific model architectures and their training are given in Appendices A \& B.

\vspace{-0.5em}
\section{Experimental Results}
\vspace{-0.5em}

\vspace{-0.25em}
\subsection{Shear Removal from Tactile Images}
\vspace{-0.5em}

We first performed an ablation study to demonstrate the separation of latent representations by the \textit{Disentangled UnshearNet} into pose and shear codes respectively. For details see Appendix D.

The first test of the performance of our approach for removing motion-induced global shear from tactile images used the multi-scale structural similarity index (SSIM)~\cite{ssim} to compare tactile images from the shear-transformed test data set with the paired canonical and the \textit{Disentangled (Entangled) UnshearNet}-generated unsheared tactile images. All three images represent the same pose. 

To compare, we computed the average SSIM between the canonical (tap)-sheared and canonical (tap)-unsheared images in several distinct scenarios (stimuli: disk edge, clover corner, teardrop corner). The unsheared images were structurally closer to the target canonical (tap) samples than their sheared counterparts. The overall SSIM value for unsheared samples was 0.93 (0.90) in comparison to 0.32 for sheared samples, with closeness to unity indicating a good match (Table~\ref{tab:ssim}). Performance was similar for all stimuli considered with \textit{Disentangled UnshearNet} achieving closer match to canonical samples than its \textit{Entangled} counterpart.




\begin{table}[ht]
\centering
\caption{\textbf{Multi-Scale Structural Similarity Index (MS-SSIM)}}
\label{tab:ssim}
\resizebox{0.9\textwidth}{!}{%
\begin{tabular}{|c|c|l|c|l|c|l|c|l|}
\hline
\multicolumn{9}{|c|}{\textbf{Image Similarity Index}}                                                                                                                                                                                                                                           \\ \hline
                                                                                                          & \multicolumn{2}{c|}{\textbf{Disk Edge}} & \multicolumn{2}{c|}{\textbf{Clover Corner}} & \multicolumn{2}{c|}{\textbf{Teardrop Corner}} & \multicolumn{2}{c|}{\textbf{Overall}} \\ \hline
\textbf{\begin{tabular}[c]{@{}c@{}}Tap Data \& Tap Data\end{tabular}}                & \multicolumn{2}{c|}{1}                  & \multicolumn{2}{c|}{1}                      & \multicolumn{2}{c|}{1}                        & \multicolumn{2}{c|}{1}                \\ \hline
\textbf{\begin{tabular}[c]{@{}c@{}}Tap Data \& Shear Data\end{tabular}}              & \multicolumn{2}{c|}{0.33}               & \multicolumn{2}{c|}{0.31}                   & \multicolumn{2}{c|}{0.33}                     & \multicolumn{2}{c|}{0.32}             \\ \hline
\textbf{\begin{tabular}[c]{@{}c@{}}Tap Data \& Unsheared (Entangled)\end{tabular}}    & \multicolumn{2}{c|}{0.91}               & \multicolumn{2}{c|}{0.90}                   & \multicolumn{2}{c|}{0.90}                     & \multicolumn{2}{c|}{0.90}             \\ \hline
\textbf{\begin{tabular}[c]{@{}c@{}}Tap Data \& Unsheared (Disentangled)\end{tabular}} & \multicolumn{2}{c|}{0.94}               & \multicolumn{2}{c|}{0.92}                   & \multicolumn{2}{c|}{0.92}                     & \multicolumn{2}{c|}{0.93}             \\ \hline
\end{tabular}%
}
\end{table}

\vspace{-0.5em}
\subsection{Reconstructed Contact Geometry}
\vspace{-0.5em}
As stated in the introduction, an application of shear removal is to aid reconstruction of the stimulus geometry from the tactile images. This is validated by using a Voronoi-based method to infer an approximate indentation field, using the method introduced in Cramphorn {\em et al}~\cite{Nathan_voronoi} on tactile data from the TacTip. The Voronoi method transforms the displacements of the tactile markers into areas of hexagonal cells tessellating the grid of the markers~\cite[Fig. 4]{Nathan_voronoi}. The change in Voronoi areas of the cells represents the skin distortion, which correlates with the indentation due to contact geometry.   

For display purposes, we fit a 3D surface to the $(x,y)$ centroid coordinates using the Voronoi cell areas as the corresponding height values (Fig.~\ref{fig:results}, top row). 
The fitted surface from the original canonical tactile images taken under tapping visibly represents the shape of the indentation on the tactile sensor, as expected (Fig.~\ref{fig:results}, top row). The same stimulus under sliding motion is visibly distorted by shear, resulting in an indentation field that does not resemble the stimulus shape (Fig.~\ref{fig:results}, top row). 

Evidently, the indentation fields reconstructed from the model-generated unsheared tactile images under sliding motion resemble those from the canonical data (Fig.~\ref{fig:results}, top row;  last column, all stimuli). Again, this confirms the successful removal of sliding-induced distortion from the tactile image. 

\vspace{-0.5em}
\subsection{Validation on Shape Exploration}
\vspace{-0.5em}
\begin{table}[b]
\vspace{-1em}
\centering
\caption{\textbf{Mean Absolute Error (MAE) in Pose Predictions}}
\label{tab:poseerr}
\resizebox{0.75\textwidth}{!}{%
\begin{tabular}{|c|c|c|c|c|c|c|c|c|}
\hline
\multicolumn{9}{|c|}{\textbf{Pose Prediction Error}}                                                                                                                                                                                                                                                                                                                                                                                                                                       \\ \hline
                                                                                 & \multicolumn{2}{c|}{\textbf{\begin{tabular}[c]{@{}c@{}}Tap Data\end{tabular}}} & \multicolumn{2}{c|}{\textbf{\begin{tabular}[c]{@{}c@{}}Sheared Data\end{tabular}}} & \multicolumn{2}{c|}{\textbf{\begin{tabular}[c]{@{}c@{}}Unsheared Data\\ (Entangled)\end{tabular}}} & \multicolumn{2}{c|}{\textbf{\begin{tabular}[c]{@{}c@{}}Unsheared Data\\ (Disentangled)\end{tabular}}} \\ \hline
\textbf{\begin{tabular}[c]{@{}c@{}}horizontal, $\tau_{x}$\\ (mm)\end{tabular}}   & \multicolumn{2}{c|}{0.43}                                                                    & \multicolumn{2}{c|}{2.72}                                                                          & \multicolumn{2}{c|}{0.95}                                                                        & \multicolumn{2}{c|}{0.64}                                                                           \\ \hline
\textbf{\begin{tabular}[c]{@{}c@{}}yaw, $\theta n_{Z}$\\ (degrees)\end{tabular}} & \multicolumn{2}{c|}{2.13}                                                                    & \multicolumn{2}{c|}{22.20}                                                                         & \multicolumn{2}{c|}{6.73}                                                                        & \multicolumn{2}{c|}{4.38}                                                                           \\ \hline
\end{tabular}%
}
\end{table}

To further confirm that our model can remove shear from transformed images, we trained a prediction network (PoseNet) on stimulus pose parameters: lateral position ($x$-horizontal, $\tau_{X}$) and in-plane orientation (yaw, $\theta n_{Z}$), using canonical tactile images from vertical tapping contacts. PoseNet performance was quantified with the mean-absolute error (MAE) between predicted pose and target when input data were canonical, sheared and model-generated unsheared tactile images respectively using both the entangled and disentangled networks (results in Table~\ref{tab:poseerr}). On the sheared data, the lowest prediction error was obtained on unsheared samples with the disentangled network, with the entangled network about 50\% higher, and on the original sheared samples about 5$\times$ poorer. Again, the \textit{UnshearNet} managed to successfully correct the distortion in sensor response cause by slide-induced global shear, with the disentangled representation giving the best performance. 

 
Another application of shear removal that also tests the real-time performance of UnshearNet is to use the above pose prediction network to servo control smoothly over varied object shapes during sliding motion. The robot successfully traced contours across all three shapes by a combination of the use of \textit{UnshearNet} for global shear removal, \textit{PoseNet} for prediction of object pose from unsheared images (Fig.~\ref{fig:results}, second row) and a simple control policy (see Appendix C). This clearly indicates that the proposed \textit{Disentangled UnshearNet} is well suited for correcting the sliding-induced shear-distorted sensor response for real-time servo control applications. 
 

\begin{figure}[t]
  \centering
  \includegraphics[width=1\textwidth]{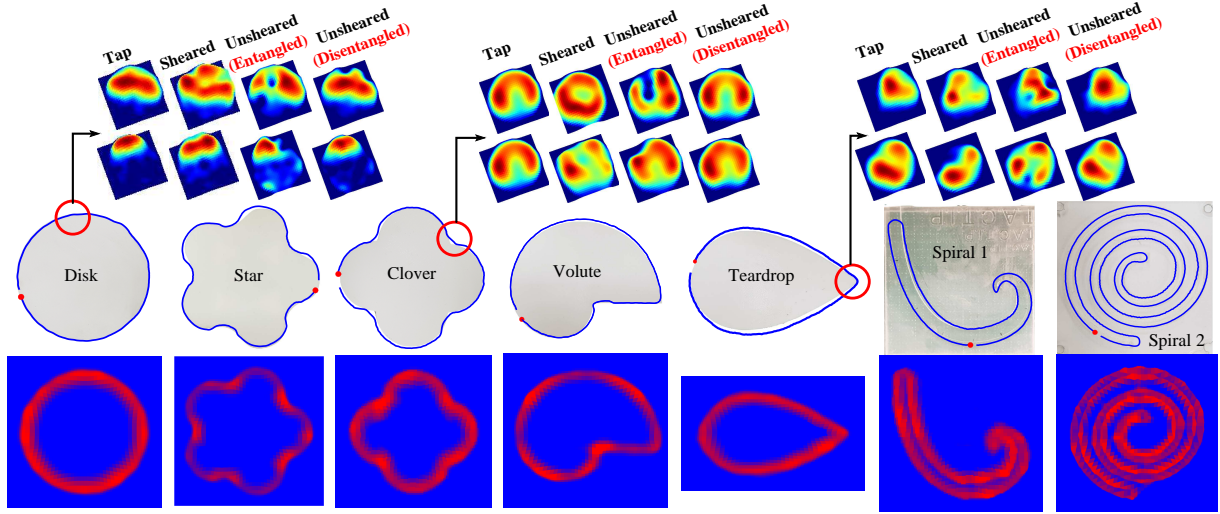}
  \vspace{-1.5em}
  \caption{\textbf{Contact geometry reconstructions, contour following and shape reconstructions.} Top row: Local contact geometry reconstruction useing $(x,y)$-centroid values and the Voronoi cell areas as \textit{z} values to fit a 3D surface using cubic interpolation. The approximate contact geometry can be recovered from sheared data using the unsheared images generated by the two types of \textit{UnshearNet} (entanged and disentangled), but not from the original sheared data. Middle row: Trajectories under robust sliding using pose estimation post removal of sliding-induced shear, overlaid on the objects, with pose estimation applied to the output of the \textit{Disentangled UnshearNet} (red dot: starting point of trajectory). Bottom row: Full object reconstructions from combining the unsheared local contact geometry reconstructions along the sliding trajectories. High-resolution figures are in Appendix E.}
  \vspace{-1em}
  \label{fig:results}
\end{figure}
\vspace{-0.5em}
\subsection{Full Object Reconstruction}
\vspace{-.5em}
As a final demonstration of our model's utility to remove shear, we reconstructed full object shapes by combined use of contact geometry estimation (section 4.3) and sliding shape exploration (section 4.4), both of which are adversely affected by motion-induced shear (Fig.~\ref{fig:results}, bottom row).

To attain full object reconstructions, we fused together the contact information extracted from unsheared images corresponding to sliding contacts (acquired during shape exploration) and linearly interpolated them over a rectangular grid (Fig.~\ref{fig:results}, bottom row). These results clearly demonstrates the effectiveness of our approach in learning the sheared-unsheared mapping that can then later be reused for multiple downstream tasks. In addition, the faithful shape reconstructions along with the shape exploration across multiple planar objects show that the model generalizes to novel situations.

\vspace{-.5em}
\section{Conclusion}
\vspace{-.5em}

In this work, we proposed a tactile image-to-image supervised convolutional neural network model to remove the effects of motion-induced global shear from the tactile sensor response while preserving the sensor deformations due to the spatial geometry of the contacted stimuli. Two \textit{UnshearNet} models were considered: one based on a basic encoder-decoder structure with an entangled pose and shear latent space, and a second that disentangled these components in the model latent space (Figs~\ref{fig:base_arch}, ~\ref{fig:prop_arch}). 


This approach was validated in three ways: (1) Comparing similarity of tactile images collected by vertically tapping without a shearing motion to those collected with an unknown shearing perturbation, for which the unsheared data passed through the \textit{Disentangled UnshearNet} gave a good match according to the structural similarity index (Table~\ref{tab:ssim}). (2) Reconstructing the indentation field from the contact geometry using sheared tactile images passed through the \textit{Disentangled UnshearNet} (Fig.~\ref{fig:results}, top row), via a (Voronoi) method used previously to reconstruct the sensor deformation~\cite{Nathan_voronoi}. The removal of motion-induced shear was apparent in the model-generated unsheared tactile images, with the contact geometry only visible after use of the UnshearNet with the original sheared image obscuring the contact shape. (3) Predicting the local pose of part of an object in a shear-insensitive manner with a prediction network trained only on canonical (tapping) data (Table~\ref{tab:poseerr}). This approach was then applied to smooth servo control around several planar objects (Fig.~\ref{fig:results}, middle row).

The combination of smooth servo control and local shape reconstruction was later used for full object reconstruction (Fig.~\ref{fig:results}, middle row), which remains an problem under active investigation in the field~\cite{suresh2020,bauza2019}. Our approach improves upon related work~\cite{Nathan_CNN_Sliding1, Nathan_CNN_Sliding2} that instead trained insensitivity to shear into the prediction network itself. An obvious drawback of this approach, even though it works well for smooth servo control, is that it does not extend readily to predict other tactile dimensions such as parameters of the local contact geometry. In that case, new labelled data would need collecting to train a new model for each intended task. Our approach, of shear removal from tactile images, instead means that a single model based on the canonical (tapping) data can be re-used.

The present study considered only planar objects in 2D and two components of pose. As recent results have shown how to extend the \textit{PoseNet} to 3D surfaces (3 pose components) or 3D edges (5 pose components) for tactile servo control~\cite{Nathan_CNN_Sliding1, Nathan_CNN_Sliding2}, we expect the present results will extend to those situations, although questions remain on the best way to do this with the large number of degrees of freedom. Although the proposed disentanglement-based approach to remove shear from sensor response has been tested on TacTip optical tactile sensor, it should hold well for other soft optical sensors. The models, however, would have to be retrained as the sensor response depends on the frictional and elastic properties of the tactile surface, which varies across sensor types.

One limitation of our current approach is that we limited our training to just translation shear. However, smooth servoing over the objects necessarily introduces rotational shear, as the servo control rotates the sensor while continuously sliding over the object surface. However, the fact that the methods performed well both in controlling the sensor and reconstructing the object geometry in the presence of this rotational shear shows that some unshearing carries over from translation to rotational shear at least for planar objects. Our expectation is that bringing rotational shear into the training will become important when extending the methods to 3D objects. Another limitation of the supervised approach investigated here is the requirement of paired canonical and sheared data, which complicates the data collection. Thus, a semi-supervised approach would be a worthwhile extension to this work, by obviating the need for paired data collection. Overall, we expect the methods developed here can can be applied more widely to various other exploration and manipulation tasks involving soft tactile sensors that are adversely affected by motion-induced shear.

\newpage
\section*{APPENDIX}
  

  

\subsubsection*{A: Network Architectures}
\paragraph{\textbf{Entangled UnshearNet:}}
The encoder \textit{E} compresses the input $256\times256$-pixel tactile image using five convolutional (Conv) layers, each followed by batch normalization (BN) and rectified linear unit (ReLU) activation layers respectively (architecture in Fig.~\ref{fig:base_arch}). The output of the last Conv layer ($8\times8\times64$) is passed to the decoder \textit{D} which upsamples it to an output $256\times256$-pixel canonical image \textit{PC}. Similarly to the Conv layers, all transposed convolutional layers (T-Conv) were followed by BN and ReLU activation layers except the output layer which used sigmoid activation function instead (see also Fig.~\ref{fig:base_arch}). 


This model was used as a baseline to evaluate the performance of proposed \textit{Disentangled UnshearNet} model on its effectiveness in removing distortion in sensor response caused by sliding-induced global shear. For details of the training, please see Appendix B.

\paragraph{\textbf{Disentangled UnshearNet:}}
The encoder has the same inputs and architecture as the \textit{Entangled UnshearNet}, except the the output of the $5^{th}$ Conv layer is followed by two additional Conv layers, one each for the two latent codes: Pose ($8\times8\times64$) and Shear ($8\times8\times64$). 


One decoder \textit{DC} takes as input pose latent code and upsamples it to $256\times256$-pixel canonical output tactile image \textit{PC}. All T-Conv layers were followed by BN and ReLU activation layers except the output layer which used sigmoid activation layer instead (architecture shown in Fig.~\ref{fig:prop_arch}). 

The other decoder \textit{DS} takes as input both pose and shear codes, merges them and upsamples to an output $256\times256$-pixel sheared image \textit{PS} (intended to match the encoder input). Apart from the above differences, \textit{DS} has the same architectures as \textit{DC}. For details of the training, please see Appendix B.



\begin{figure*}[ht]
  \centering
  \includegraphics[width=0.65\textwidth]{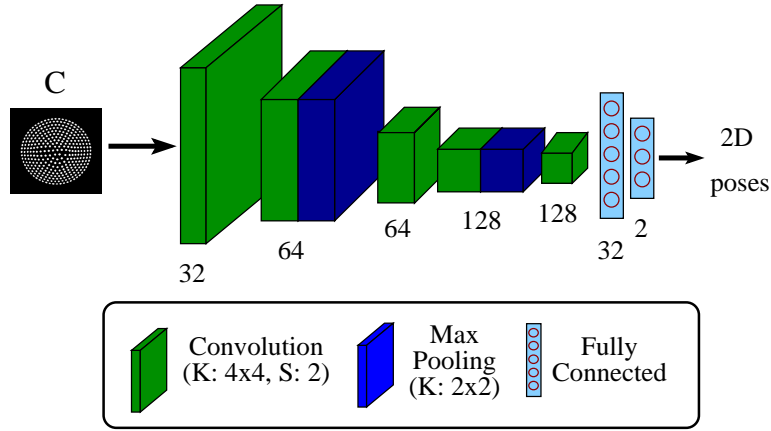}
  \caption{\textbf{PoseNet schematic:} This network takes as input, a tapping binary sensor image, \textit{C}. The convolution part compresses input to extract relevant features which are then combined by fully connected part to predict continuous-valued object pose parameters at output.}
  
  \label{fig:posenet}
\end{figure*}

\paragraph{\textbf{PoseNet:}}
This model takes as input $256\times256$ image, compresses it to extract features using the Conv part which are then combined using fully connected (FC) part to predict continuous-value pose components at the output. In total, the network consists of five convolution layers, two max pooling (MP) layers following $2^{\rm nd}$ and $4^{\rm th}$ Conv layers; the CONV layers were followed by BN and ReLU activation layers respectively. The $1^{\rm st}$ FC layer used ReLU activation while the output FC layer used sigmoid activation. For details of the architecture and training see Fig.~\ref{fig:posenet} and Appendix B. 

\newpage
\subsection*{B: Training Details}
All the inputs (binary $256\times256$-pixel images) and outputs (binary $256\times256$-pixel images for image-to-image models, pose parameters for image-to-pose model) were scaled to the range $[0, 1]$. All networks weights were initialized from a zero-centered normal distribution with 0.02 standard deviation. Both image-to-image models were trained on data collected from all three stimuli shapes (Fig.~\ref{fig:sensor} (c)) as discussed in section 3.2.

For image-to-image models, we used a batch size of 32. All convolutional/transposed convolutional layers used L1/L2 regularization (10e-4) along with random image shifts, 1-2\% of image size to prevent overfitting. The \textit{L2 loss} computed across the entire image and \textit{L1 patch loss} -- computed between 100 random crops of size $20\times20$ of generated unsheared images and corresponding canonical images -- were used to train the networks in the ratio 10:1.  In case of \textit{Disentangled UnshearNet}, the encoder (E) and shear decoder (DS) were trained using reconstruction loss.

For image-to-pose model, we used a batch size of 256. Like image-to-image models, all layers used L1/L2 regularization (10e-4) along with random image shifts, 1-2\% of image size to prevent overfitting. The network was trained via the \textit{L2 loss} computed between predicted and target pose parameters. 

We used ADAM optimizer~\cite{Kingma_AdamOpt} with learning rate of 0.0001,  $\beta1$ = 0.5 and $\beta2$ = 0.999. We used learning rate scheduling for training. For image-to-image models, the learning rate was reduced to one-fifth and one-tenth after 40 and 80 epochs. The models were trained, in total, for 100 epochs. For image-to-pose model, the learning rate was reduced to half and one-tenth after 100 and 200 epochs respectively. The model was trained for 250 epochs. For all models, the model with best validation accuracy was used for testing.

Finally, training and optimization of the networks was implemented in the Tensorflow 2.0 library on a NVIDIA GTX 1660 (6 GB memory) hosted on a Ubuntu machine. 

\subsection*{C: Control Policy for Continuous Contour Following}
Local pose estimation allows a robot to maintain contact while safely moving over the object, thus enabling complex robot-object interactions. To demonstrate continuous 2D contour following, we used a simple control policy with following two aims: 1) keep the sensor normal to object surface while in motion and 2) at every time step \textit{t}, move the sensor tangentially along the surface by a predefined step (0.5 mm in this case). To achieve these aims, a discrete-time proportional-integral (PI) controller was implemented to output a change in the pose of the sensor ($\Delta p(t)$) in its reference frame
\begin{equation*}
    \Delta p(t) = K_{p}e(t) + K_{i}\sum_{t'=0}^{t} e(t')
\end{equation*}

where $K_{p}$ and $K_{i}$ are diagonal gain matrices with proportional gain of 0.5 and integral gains of 0.3 and 0.1 for translations and rotations respectively. $\textit{e}$ was error between the predicted pose and a reference normal to the edge.

\newpage
\subsection*{D: Disentanglement of Latent Representations}
An ablation study was used to verify the separation of latent representations in the \textit{Disentangled UnshearNet} into pose and shear codes respectively. To do this, we passed the `Shear Code' to the unsheared reconstruction decoder (DC) instead of the ‘Pose Code’. As expected, this led to a severe degradation in performance with mean-square error (MSE) between the ground truth images \textit{C} (tap) and unsheared images \textit{PC} increasing to 0.22, an order of magnitude higher than original of 0.023. In similar fashion, the SSIM index dropped to $2\%$ when using the Shear Code to reconstruct the unsheared images \textit{PC} instead of the original $93\%$ when the Pose Code was used. Likewise, asimilar degradation was observed when the `Shear Code' was replaced by the Pose Code to reconstruct the sheared input \textit{S}. The mean-squared error between \textit{S} and the sheared output \textit{PS} increased to 0.1, which was 50-times the original of 0.002 when both `Pose and Shear Codes were used to reconstruct \textit{PS}. The SSIM index showed a similar trend with the similarity dropping from $99.5\%$ to  only $11\%$. This shows that the Shear Code is indeed encoding relevant information required for successful reconstruction of sheared input \textit{S}. These results clearly demonstrate that the \textit{Disentangled UnshearNet} successfully disentangles the latent representations as desired.

\newpage
\subsection*{E: High Resolution Results}
\vspace{0cm}
\begin{figure*}[ht]
  \centering
  \includegraphics[width=0.5\textwidth]{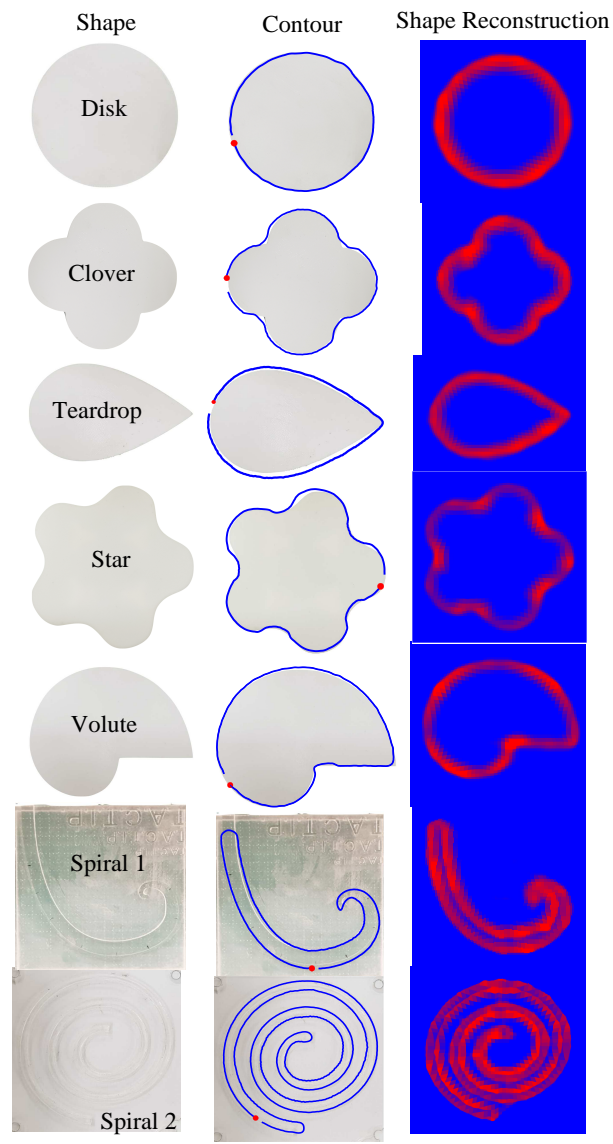}
  \vspace{-0.25cm}
  \caption{\textbf{Contour following \& Shape Reconstruction.} Left: Stimuli shapes, Middle: Robust sliding across shapes in left row post removal of shear using \textit{Disentangled UnshearNet}. Red dot shows the starting point/initial contact. Right: Full shape reconstruction post removal of shear using \textit{Disentangled UnshearNet.}}
  
  \label{fig:highres_results}
\end{figure*}

\end{document}